\documentclass[letterpaper,twocolumn,fleqn]{article} 

\usepackage{ist}
\usepackage{amssymb}
\usepackage{color}
\usepackage{algorithm}
\usepackage{algorithmic}
\usepackage{array}
\usepackage{tikz}
\usetikzlibrary{shapes}
\usepackage{amsmath}
\usepackage{url}
\usepackage{graphicx}
\usepackage{tabularx}
\usepackage[stable]{footmisc}
\usepackage{xcolor}
\usepackage{caption}
\usepackage{hyperref}
\usepackage{subcaption}

\def\##1{\relax\ifmmode\mathchoice
{\mbox{\boldmath$\displaystyle#1$}}
{\mbox{\boldmath$\textstyle#1$}}
{\mbox{\boldmath$\scriptstyle#1$}}
{\mbox{\boldmath$\scriptscriptstyle#1$}}\else
\hbox{\boldmath$\textstyle$}\fi}

\title{VRUNet: Multi-Task Learning Model for Intent Prediction of Vulnerable Road Users}

\author{Adithya Ranga$^{1}$, Filippo Giruzzi$^{2}$, Jagdish Bhanushali$^{1}$, Emilie Wirbel$^{3}$, Patrick P\'erez$^{4}$, Tuan-Hung Vu$^{4}$, Xavier Perotton$^{3}$\\
$^{1}$Valeo Driving Assistance Research, San Mateo, CA, USA \\
$^{2}$MINES ParisTech, Paris, France \\
$^{3}$Valeo Driving Assistance Research, Paris, France \\
$^{4}$Valeo.ai, Paris, France
}

\begin{document} 

\maketitle 



\begin{abstract}
Advanced perception and path planning are at the core for any self-driving vehicle. Autonomous vehicles need to understand the scene and intentions of other road users for safe motion planning. For urban use cases it is very important to perceive and predict the intentions of pedestrians, cyclists, scooters, \textit{etc.}, classified as vulnerable road users (VRU). Intent is a combination of pedestrian activities and long term trajectories defining their future motion. In this paper we propose a multi-task learning model to predict pedestrian actions, crossing intent and forecast their future path from video sequences. We have trained the model on naturalistic driving open-source JAAD \cite{kotseruba2016joint} dataset, which is rich in behavioral annotations and real world scenarios. Experimental results show state-of-the-art performance on JAAD dataset and how we can benefit from jointly learning and predicting actions and trajectories using 2D human pose features and scene context.
\end{abstract}


\section{INTRODUCTION}\label{sec:introduction} 
\begin{figure*}[tb]
\centering
\includegraphics[width=0.95\linewidth]{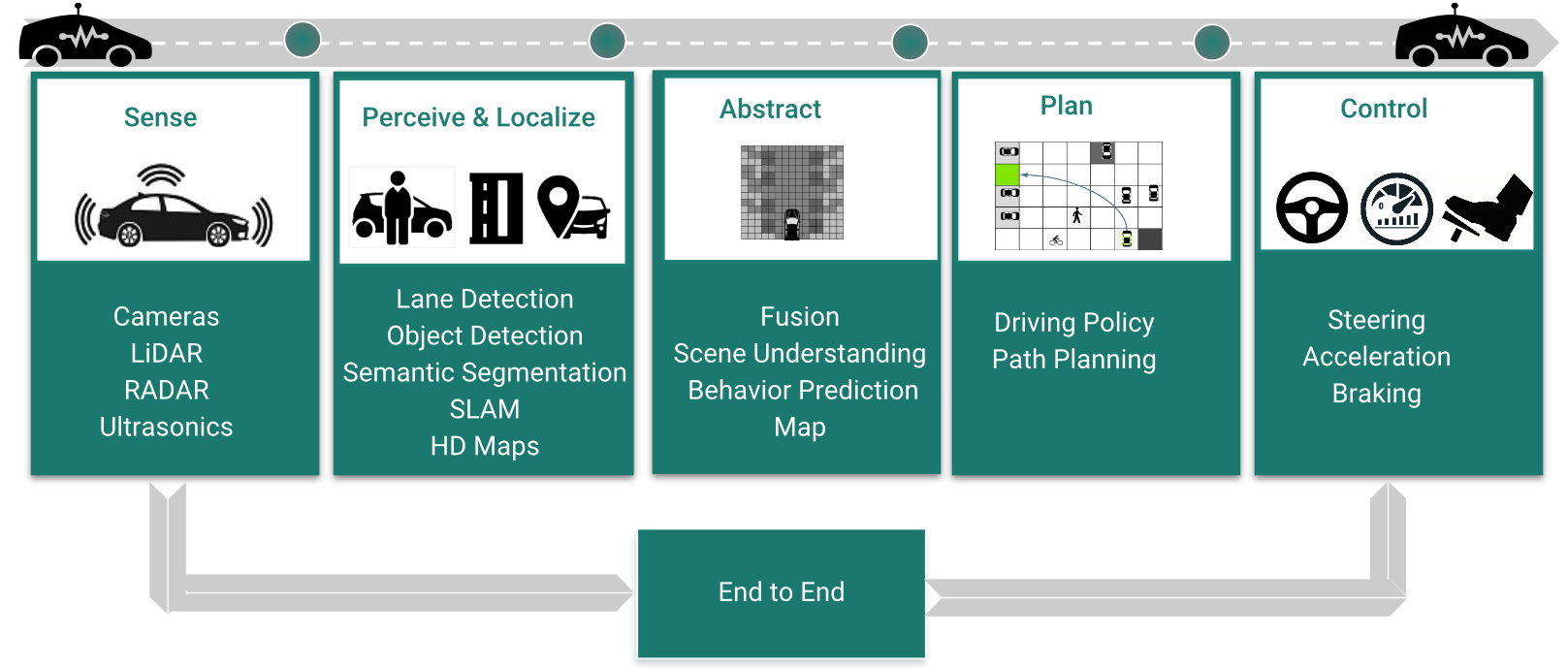}
\caption{Modules in an autonomous driving pipeline. 
}
\label{fig:ADBlockDiagram}
\end{figure*}

With the advancements in artificial intelligence (AI) and deep learning we are able to tackle some of the challenging problems, in the field of autonomous systems and robotics. Automated Driving (AD) or self-driving vehicles are becoming more common on urban and highway roads, and are able to handle many complex driving scenarios. Figure \ref{fig:ADBlockDiagram} illustrates the entire AD architecture starting from sensing, all the way to lateral and longitudinal control of the vehicle. Advanced perception including object detection and scene understanding, followed by detailed abstraction of information specific to individual objects in the scene, are crucial for path planning. 

\begin{figure}[tb]
\centering
\includegraphics[width=\linewidth]{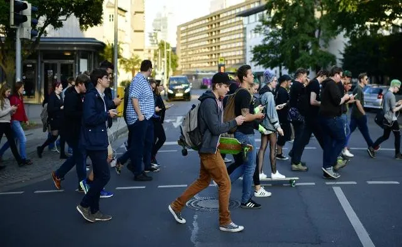}
\caption{Distracted pedestrians while walking/crossing on urban roads 
}
\label{fig:PedestriansGeneral}
\end{figure}

One of the key challenges with path planning for automated driving on urban roads is that the vehicles have to constantly interact with pedestrians, cyclists, scooters etc. generally identified as Vulnerable road user's (VRU). VRU's on road move with specific goals in mind, while respecting certain rules and also directly interacting with the other actors in scene. It is increasingly apparent these days, that many pedestrians are distracted using devices (eg. cell phones, headset's) as seen in Figure \ref{fig:PedestriansGeneral}, and put themselves and the traffic around at higher risk. This survey by Rasouli et al.\cite{rasouli2018autonomous} discusses all the factors to be considered for understanding VRU's and their interactions. Human cognition is very good at understanding and anticipating their actions and intentions on road. To achieve naturalistic driving behavior and safely interacting with other road user's, AI needs to match these levels of human intelligence. This problem of activity recognition and prediction, is gaining significant attention from the AI and computer vision community, especially focused towards applications of automated driving and surveillance. 

Most of the earlier research methods propose a model based approach or probabilistic solution to predict the intent of VRU's in real time. More recently, with the availability of open-source datasets for action recognition and automated driving, many machine learning/deep learning models are being developed. Most of these approaches focus on activity recognition or address intent as a time-series prediction problem separately. The overall intention of actors on road can be better summarized from their short term discrete actions and continuous forecasting of their future positions. The actions performed by pedestrians in the past can be used to predict their future actions and how they move on the road. Pedestrians in the scene also navigate respecting some defined rules of the scene and interacting with other objects in the scene, like walking on the cross-walk while crossing, stopping for traffic at traffic lights, yielding for other vehicles, etc. To this end, we propose a single multi-task learning model that jointly predicts the actions, crossing intent and also trajectories from video sequences. Firstly, to better understand the actions of the persons, we abstract low level features regarding their body pose and efficiently track them in the scene. Secondly for intent prediction we use scene semantics as an additional input to the multi-task prediction model. Experiments on JAAD \cite{kotseruba2016joint} dataset show better results than the baseline, for pedestrian intent using this multi-task learning approach.

The rest of the paper is organized as follows: a review of the related work is presented in Section \ref{sec:related}. Details regarding our research methodology and network architecture are discussed in Section \ref{sec:Methodology}. Experiments conducted on JAAD and the results are illustrated in Section \ref{sec:Experiments}. Section \ref{sec:Conclusions} concludes the paper. 

\section{RELATED WORK}\label{sec:related}

\subsubsection{Activity Recognition and prediction} 
Human activity recognition and prediction from video data has been studied for some time now. Many works have been proposed using Recurrent Neural Networks (RNN). \cite{Singh_2016_CVPR}, \cite{Du_2015_CVPR} and \cite{veeriah2015differential} are some of the recent works on action recognition from videos using RNN's for single individuals.\cite{Ma_2016_CVPR} proposes using LSTM's for early activity detection. Also there are some recent methods \cite{ibrahim2015hierarchical}, \cite{deng2015structure} for action recognition for groups and \cite{Bagautdinov_2017_CVPR} discusses an end-to-end approach of detecting persons and jointly predicting their behaviors. 

\subsubsection{Trajectory Prediction} 
Predicting the future path of pedestrians in videos is a well known area of research. Most of the models are either model based, probabilistic models or more recently - deep learning based approaches. The authors in \cite{schller2019constant} compare constant velocity models (CVM) with probabilistic and deep learning approaches for trajectory prediction. In \cite{Rehder_2015_ICCV_Workshops}, a method for improving the performance of model based prediction using goal or final destination of pedestrians as latent variable is proposed. Well known deep learning approaches use RNN's like LSTM encoder-decoder architecture trajectory prediction like \cite{Shi_2019}. More recent deep learning approaches focus on improving the trajectories considering human social interaction in crowds like the work done in Social-LSTM \cite{Alahi_2016_CVPR}. Social-GAN \cite{gupta2018social} used adversarial training and SR-LSTM \cite{zhang2019srlstm} added a refinement module to LSTM network using neighbors intentions, aiming at improving the accuracy of trajectory performance in crowded scenes. Also Scene-LSTM \cite{manh2018scenelstm} uses scene data along with pedestrian data and a grid structure to learn and predict human trajectories with reduced location displacement errors. Most of these deep learning and socially aware approaches are derived on static cameras looking at crowded scenes without considering the current or past actions or behaviors of the persons.

\subsubsection{Intent Prediction for Automated driving}

Recently, with increased attention towards pedestrian safety and AD, research activities \cite{rasouli2018autonomous} are focusing more on behavioral science of pedestrians and their interactions with other road users. Many intention prediction models have been proposed so far especially for pedestrians.

\textbf{Model based approaches} like \cite{Schneider2013PedestrianPP} and \cite{Keller2014WillTP} rely on dynamic models and probabilistic estimation techniques. In some recent applications like \cite{Kooij2014ContextBasedPP} pedestrian context like awareness when looking at the vehicle and metrics like distance to the curb were taken into consideration to improve the accuracy of intention prediction with dynamic models. In \cite{Madrigal2014IntentionAwareMP} social forces or interactions were added to the dynamic models to further improve the accuracy of model based intention prediction. Most of the model based work is done with staged scenarios, while treating the pedestrian as any other point in space and not including any scene context.

\textbf{Machine learning (ML) approaches} in  \cite{Koehler2012} and \cite{Koehler2013} use SVM and pose features to predict the crossing intent. To further improve the performance, in \cite{Rasouli_2017_ICCV} and \cite{Schneemann}, \textbf{scene context} like existence of a traffic light, cross walks and lane signature were included for a sequence of frames. More recently researchers in \cite{fang2019intention} have used image sequences and skeleton-based features for pedestrians and cyclists and predicted crossing/not-crossing intent values. 

\textbf{Deep learning approaches} in \cite{Rasouli_2017_ICCV} are used to predict the walking/standing and looking/not-looking state of pedestrians separately from cropped images. Intent as a time-series trajectory prediction problem is discussed in \cite{Saleh} and a stacked LSTM model is used to predict the future positions of VRU's without considering their 2D pose, behaviors and scene context. In \cite{SalehMTL} researchers propose a multi-tasking model to predict the pose orientation and standing/walking behavior of pedestrians from images. More recently work done by researchers in \cite{liang2019peeking} is very interesting as they jointly train a multi-tasking model to predict activities and trajectories as an auxiliary task. The work is focused on a static camera configuration overlooking crowded scenarios, whereas our approach is focused on naturalistic driving scenarios for automated driving.

\section{METHODOLOGY} \label{sec:Methodology} 
In urban scenarios, history of pedestrians behaviors like gait i.e. if someone is walking/standing, awareness levels, orientation, distraction and social interactions, influence the future state or actions. Also the future location or goal of the person in the scene can be determined by their past actions and correlating it with scene semantics. Motivated by this, we developed a single multi-tasking model that predicts the behaviors, crossing intentions and future trajectories of VRU's in the scene. 

\begin{figure*}[tb]
\centering
\includegraphics[width=0.95\linewidth, height=4cm]{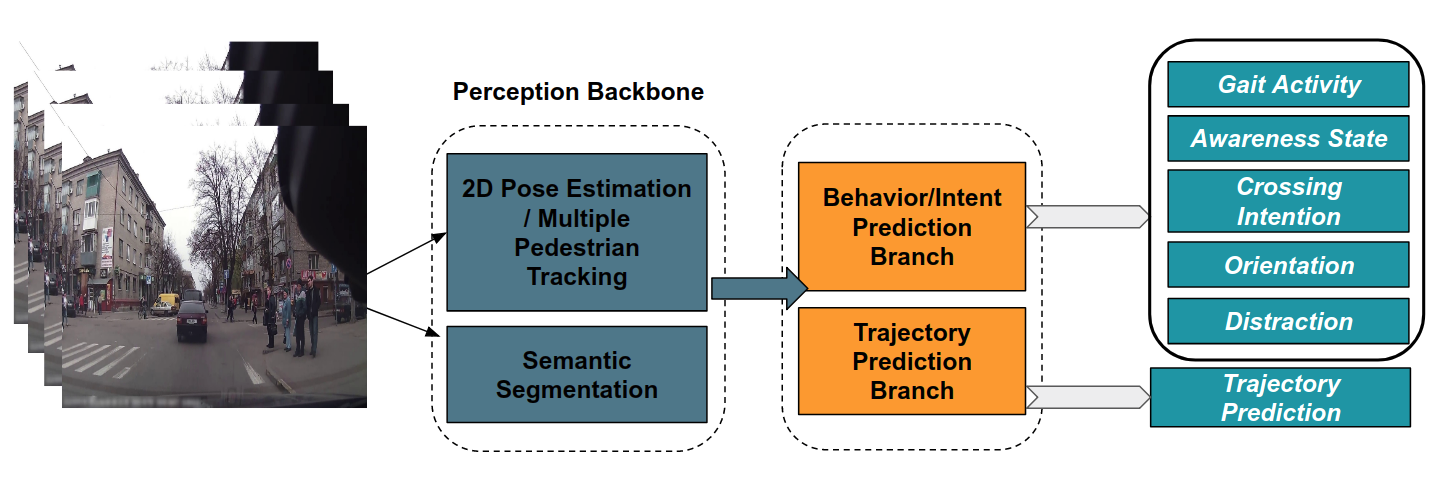}
\caption{Overview of the our research approach. Video sequences are processed through perception backbone to generate tracked object (2D pose and bounding box) and scene (segmentation mask) context. Joint model is trained on the perception outputs to predict actions, crossing intent and trajectory in the videos.
}
\label{fig:blockdiagram}
\end{figure*}

Our system as shown in Figure \ref{fig:blockdiagram} processes sequences of frames from camera first through the perception backbone, to obtain 2D pose or skeleton for all the persons in the scene and track them throughout the sequence. Also we process each frame to extract scene semantics using a semantic segmentation model. Given the 2D pose, bounding box and scene context from time 1 to $T_{current}$, the model classifies the current state or action at $T_{current}$ for each person, and predicts the future crossing/not-crossing intent at time $T_{current+horizon}$. The model also simultaneously predicts the future positions in the image i.e. the trajectory from time $T_{current}$ to $T_{current+horizon}$. Here \textit{horizon} is the defined prediction duration or look ahead into the future.

\subsection{Dataset} \label{subsec:Dataset}
One of the main challenges for deep learning based intent prediction for AD, is creating behavioral annotations for VRU's from natural driving data. This problem is mostly addressed by JAAD dataset \cite{kotseruba2016joint}, that contains rich behavioral labels for persons that interact with the driver and also provides possible intent values for specific persons in the scene. The dataset also provides additional contextual labels for each person and high level scene annotations are available. Dataset doesnt include ego-vehicle odometry information currently.

Additionally, we leverage COCO \cite{lin2014microsoft} and PoseTrack datasets \cite{PoseTrack} for fine-tuning the 2D pose detection backbone network. For scene context, a semantic segmentation network is pre-trained on Cityscapes dataset \cite{cordts2016cityscapes} and is used to obtain the encoded scene masks for our model.

\subsection{Formal definition of Behaviors and Intention} \label{subsec:definition}
In our model we focus on the following actions and crossing intent of VRU's:
\begin{itemize}
    \item \textbf{Gait}: If the person in scene is "Walking or Standing"
    \item \textbf{Attention}: If the person is directly looking at the vehicle or not (Looking / Not Looking)
    \item \textbf{Orientation}: the pose orientation of the person with respect to the viewing angle (Left / Right / Front / Back)
    \item \textbf{Distraction}: If the person is distracted with a phone (Phoning / Not Phoning)
    \item \textbf{Crossing Intention}: This tells us if the person will be crossing or not the road/lane in front of the vehicle at time $T_{current+horizon}$, where \textit{"horizon"} is the prediction period in seconds (Crossing / Not Crossing)
\end{itemize}
We process the JAAD data to refine annotated labels to the above described class labels for training and testing the model performance. 

\subsection{Perception Backbone - Pose Estimation and Scene Understanding} \label{subsec:perception}
\begin{figure}[h]
\centering
\begin{subfigure}{0.5\linewidth}
    \includegraphics[width=\linewidth, height=4cm]{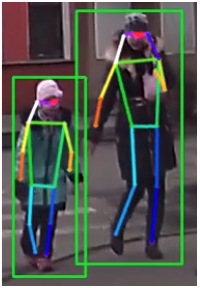}
    \caption{child and adult}
    \label{fig:pose_a}
\end{subfigure}%
\begin{subfigure}{0.5\linewidth}
    \includegraphics[width=\linewidth, height=4cm]{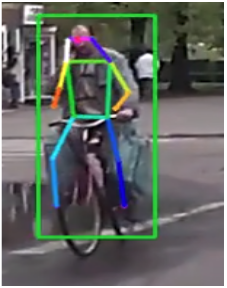}
    \caption{cyclist}
    \label{fig:pose_b}
\end{subfigure}%
\newline
\newline
\newline
\begin{subfigure}{0.5\linewidth}
    \includegraphics[width=\linewidth, height=4cm]{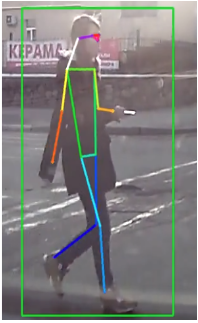}
    \caption{adult using phone}
    \label{fig:pose_c}
\end{subfigure}%
\begin{subfigure}{0.5\linewidth}
    \includegraphics[width=\linewidth, height=4cm]{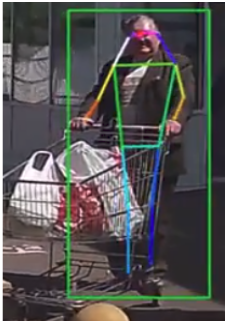}
    \caption{adult with cart}
    \label{fig:pose_d}
\end{subfigure}
\newline
\newline
\caption{2D human pose prediction visualization on JAAD data}
\label{fig:pose}
\end{figure}

In many previous approaches for activity recognition and prediction, persons in the scene are represented as any other points or objects in space as bounding boxes. Deep networks are trained on the full image, video sequences or cropped bounding boxes to recognize the actions. However most of the high level actions for humans can be abstracted from their skeleton or pose information. By representing the person as a 2D skeleton most of the dynamics could be accurately captured and the performance could greatly improve using less denser networks. To this end a pre-trained pose estimation network, \textit{PifPaf} \cite{kreiss2019pifpaf} trained on COCO dataset \cite{lin2014microsoft} and fine-tuned on Posetrack \cite{PoseTrack}, is used to obtain the keypoints and object bounding boxes. There are a total of 17 keypoints - $(X_i, Y_i, V_i)$, detected for each person at any given instance where $X_i$ and $Y_i$ are the pixel locations and $V_i$ is the visibility score of keypoint $i$. Figure \ref{fig:pose} shows the visualization of 2D pose prediction and boxes for sample persons as seen from JAAD data.

Every person in the scene needs to be tracked through the sequence, to model the temporal changes by observing the change in their features. To track multiple pedestrians in the scene we use the research approaches from \cite{Wojke2017simple} and \cite{Wojke2018deep}. This tracking technique uses the measured state of the detected object in the scene and a Kalman filter to track and update it over time. There is an additional CNN model trained on large scale person re-identification dataset \cite{Zheng2016MARSAV}, where the CNN features are associated using a defined metric to improve tracking performance. Each person in the scene is tracked and the 2D pose, boxes are extracted for the track. For a track or sequence length of \textit{N}, each person in the scene will have keypoints or pose features of dimension $N\times 17\times 3$ and bounding boxes of dimension $N\times 4$.

The perception backbone also includes a semantic segmentation module with a VGG16 encoder and UNET \cite{ronneberger2015unet} decoder architecture that is pre-trained on cityscapes dataset. This module associates all the pixels of the scene with their respective classes and gives us a full scene understanding. For an input image resolution of $W\times H\times 3$ the segmentation mask output is of the same resolution  where each pixel location has the class index. Hence the dimension of the scene context mask for a sequence length \textit{N} is $N\times W\times H\times 1$. Given the model was not trained on JAAD we see domain gap where the performance is lower for some classes.

\begin{figure}[h]
\centering
\begin{subfigure}{\linewidth}
    \includegraphics[width=\linewidth]{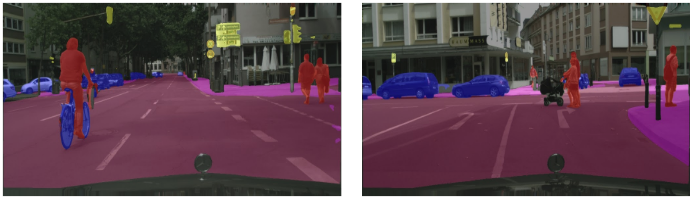}
    \caption{segmentation visualization on cityscapes test set}
    \label{fig:segmentation_a}
\end{subfigure}
\newline
\newline
\newline
\begin{subfigure}{\linewidth}
    \includegraphics[width=\linewidth]{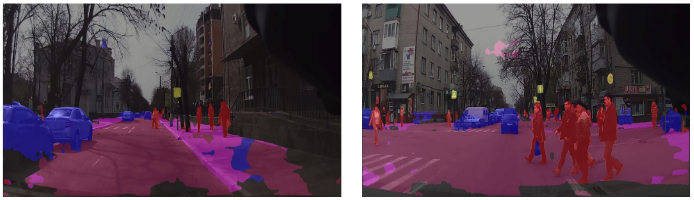}
    \caption{segmentation visualization on JAAD data}
    \label{fig:segmentation_b}
\end{subfigure}
\newline
\newline
\caption{semantic segmentation visualization on cityscapes (top) and JAAD data (bottom)}
\label{fig:segmentation}
\end{figure}

\subsection{Intent and Trajectory Prediction} \label{subsec:models}

\subsubsection{Modular approach}
\begin{figure*}[tb]
\centering
\includegraphics[width=\linewidth, height=7cm]{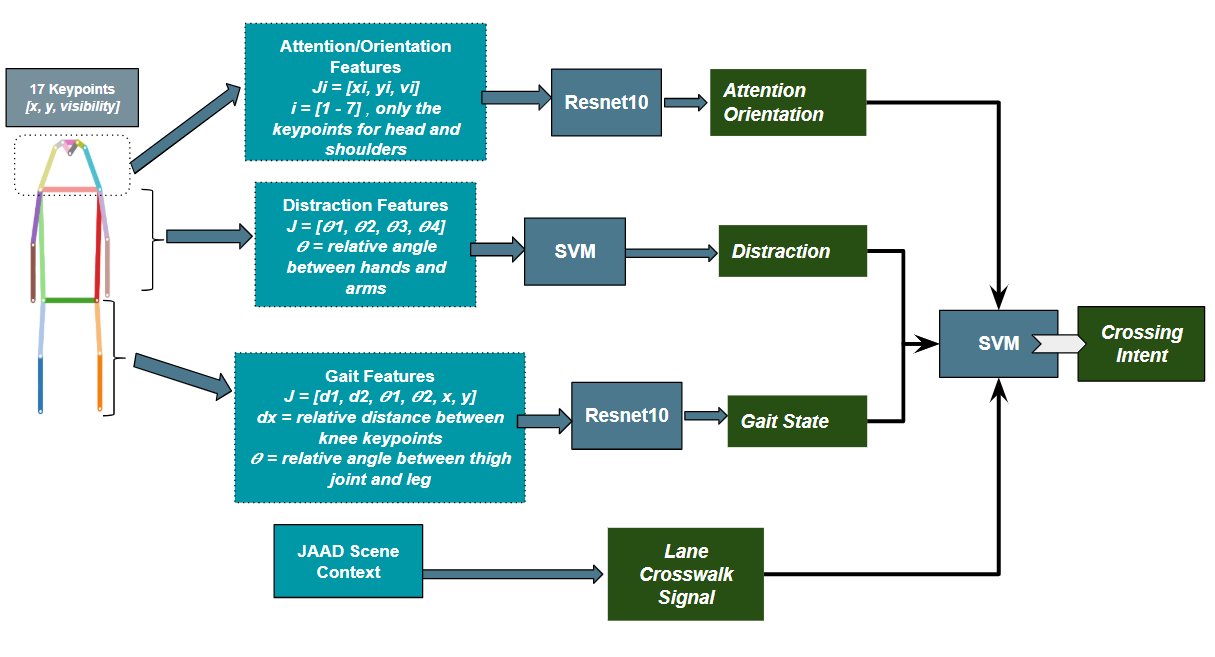}
\caption{Modular approach showing specific features and models used for each task
}
\label{fig:BenchmarkModular}
\end{figure*}

As a part of this research, we have trained separate models for each task as shown in Figure \ref{fig:BenchmarkModular}. This work is done to establish a baseline for each task separately for later comparison. We use 2D human pose and pre-calculate the geometrical features like angles and distances between joints and keypoints. The following mentioned models are separately trained with specific features to predict the behaviors (gait, attention, distraction), orientation and overall crossing intent.

\begin{itemize}
    \item \textbf{Gait Model:} As mentioned in \ref{sec:Methodology}, for gait we predict if the person in walking or standing in the scene. To determine the gait state of persons we use the keypoints for legs (knees and ankles). We calculate the features $(d_1, d_2, \theta_1, \theta_2, x, y)$, where $d_1$ and $d_2$ are respectively the distances between right/left ankles and knees, $\theta_1$ and $\theta_2$ are angles between the limbs and $x$ and $y$ is the hip center of the person. We stack the features for a sequence observation length of \textit{N}, so that the temporal change in features is captured well. A 1D Resnet-10 model is trained for the binary classification task with input shape $Batch\times (N\times 6)$ and to optimize a cross-entropy loss function.
    \item \textbf{Attention + Orientation:} For this task we mainly focus on the keypoints of the upper body i.e. head -- eyes, nose, ears, and shoulders. A total of 7 keypoint values i.e. $(x_i, y_i, v_i)$ for the stacked last $N$ frames are used as input [input shape - $Batch\times (N\times 21)$], to train a separate 1D Resnet-10 model. The model predicts both attention (Looking or Not Looking) and orientation (Left, Right, Front or Back) simultaneously, and is trained with a weighted cross-entropy loss.
    \item \textbf{Distraction:} This is a binary classification task to determine if the VRU is phoning. Angles $(\theta_{l}, \theta_{r}, \theta_{lr, hands}, \theta_{lr, upper})$, where $\theta_{l}$ and $\theta_{r}$ are the angles between lower arm and upper bicep for each hand, and $\theta_{lr, hands}, \theta_{lr, upper}$ are the angles between left and right hands and upper bicep respectively. These pre-calculated features are stacked over the last N frames and we treat it as a binary classification task using a support vector classifier (SVC) with radial basis function (RBF) kernel.
    \item \textbf{Crossing Intent:} The main goal here is to predict if the person in the scene will cross or not in front of the vehicle at some defined future time $T_{current + horizon}$, using persons context and scene information from time 1 to $T_{current}$ as input. For training this model we use the past $N$ frames and generate all the behavior states (gait, attention, distraction) and orientation values from the pre-train action recognition models. Additionally, we do binary encoding of scene context annotations from JAAD for the presence of (traffic light/sign, cross walk, lane width(narrow/wide)). Given this we have a $Batch\times (N\times 9)$ sized input with 9 features. A simple support vector model is fitted, to classify whether the person will cross the road in front or not.
\end{itemize}

\subsection{VRUNet model architecture}
\begin{figure*}[tb]
\centering
\includegraphics[width=0.95\linewidth, height=9cm]{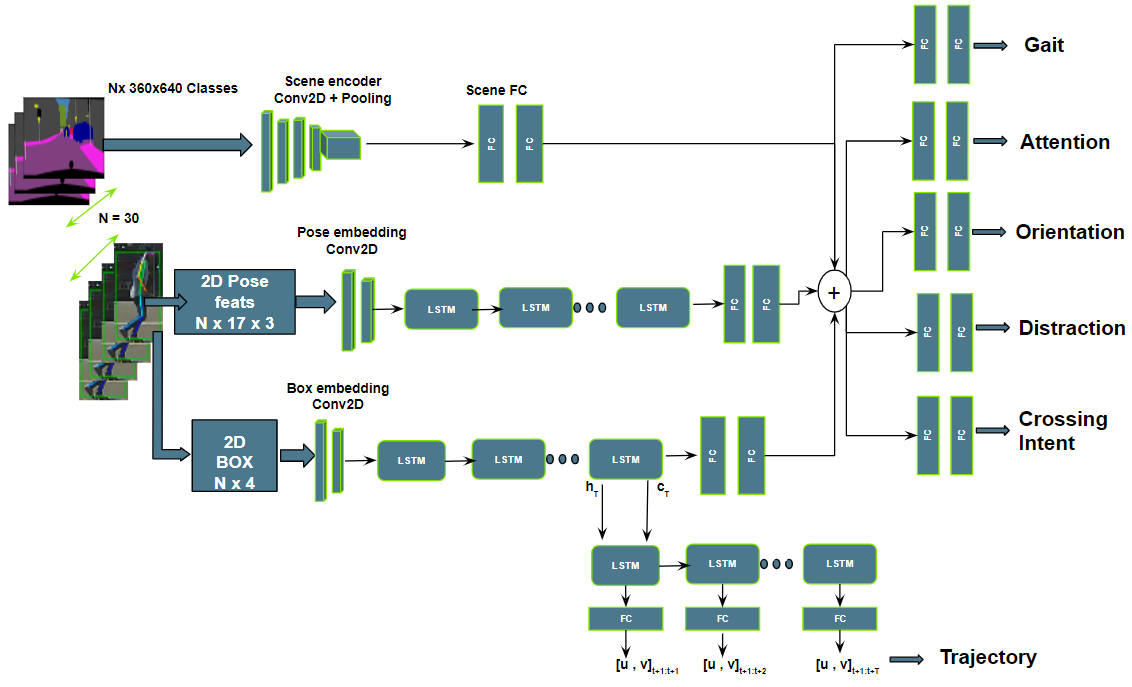}
\caption{VRUNet Multi-task Model Network Architecture
}
\label{fig:VRUNet}
\end{figure*}

Intent of all the persons in scene has strong temporal dependency with the past actions, how they navigate the scene while interacting with other actors. To this end, we propose an end-to-end trained multi-task model as shown in Figure \ref{fig:VRUNet}. Using this multi-tasking approach we jointly predict the actions, crossing intent and trajectory of VRU's from video sequences. Multi-tasking reduces the overall compute and memory requirements, by weight sharing. Inputs to this model are 2D pose features, object bounding boxes and scene semantic masks that are processed from the perception backbone. Using the object context and scene context from time 1 to $T_{current}$, we classify actions(gait, attention, distraction) and orientation of the person and predict crossing intent at time $T_{current+horizon}$. Also we jointly predict the positions of the VRU from time $T_{current}$ to $T_{current+horizon}$ in image coordinates. Input sizes to the network and the resolution of scene mask are as defined in Table \ref{tab:VRUNet-input}.

\begin{table}[h]
\begin{center}
\caption{Input Layers for VRUNet}
\label{tab:VRUNet-input}
\begin{tabular}{r|ccc}
Name & Size \\
\hline
Input Image & $N\times 360\times 640\times 3$ \\
Input Scene Mask & $N\times 360\times 640\times 5$ \\
Input Box & $N\times 4$ \\
Input Pose & $N\times 17\times 3$ \\
\end{tabular}
\end{center}
\end{table}

Input size of pose features is $Batch\times N\times 17\times 3$ where $Batch$ will be the maximum number of VRU instances we train in any given sequence. Pose input is first processed through a embedding layer as shown in Table \ref{tab:VRUNet-Pose} comprising of 2D convolutions. The output of the pose embedding layer is then input to a stacked LSTM encoder and then finally processed through fully connected layers. Similarly the input size for bounding box features is $Batch\times N\times 4$ and is processed through embedding layers followed by stacked LSTM architecture as defined in Table \ref{tab:VRUNet-Pose}. 

\begin{table}[h]
\begin{center}
\caption{Pose and Bounding Box Encoding Layers for VRUNet}
\label{tab:VRUNet-Pose}
\begin{tabular}{r|cc}
Name & Size $\times$ Filters & Stride\\
\hline
Conv \#1 & $3\times 3\times 256$ & 2 \\
Conv \#2 & $3\times 3\times 256$ & 2 \\
LSTM \#1 & 256 & N/A \\
FC \#1 & 256 & N/A \\
FC \#2 & 256 & N/A \\
\end{tabular}
\end{center}
\end{table}

Scene semantic segmentation mask from the perception backbone has 5 classes (road, car, pedestrian, sidewalk, traffic sign) that we use for this model. The segmentation mask is then binary encoded to produce sematic features of shape $H\times W\times classes$, where classes is 5. Semantic segmentation model outputs the mask with a resolution of $512\times1024$. This is then reshaped to a resolution of $360\times 640$, hence the shape of input scene features after binary encoding for the sequence is $Batch\times N\times 360\times 640\times 5$. This input is then used to compute a mean mask along the time axis before processing through the model. It is then encoded using 2D convolution and max pooling layers followed by fully connected layers as shown in Table \ref{tab:VRUNet-Scene}.

\begin{table}[h]
\begin{center}
\caption{Scene Encoding Layers for VRUNet}
\label{tab:VRUNet-Scene}
\begin{tabular}{r|ccc}
Name & Size & Filters number & Stride \\
\hline
Conv \#1 & $3\times 3$ & 256 & 2 \\
Conv \#2 & $3\times 3$ & 256 & 2 \\
Maxpool \#1 & $2\times 2$ & N/A & 1 \\
Conv \#3 & $3\times 3$ & 512 & 2 \\
Conv \#4 & $3\times 3$ & 512 & 2 \\
Maxpool \#1 & $2\times 2$ & N/A & 1 \\
FC \#1 & 1024 & N/A & N/A \\
FC \#2 & 1024 & N/A & N/A \\
FC \#3 & 256 & N/A & N/A \\
FC \#4 & 256 & N/A & N/A \\
\end{tabular}
\end{center}
\end{table}

Outputs from scene, pose and bounding box encoding branches are fused channel-wise and processed through separate branches with fully connected layers to predict the actions and crossing intent. Given the sequence of encoded poses and bounding boxes, the model outputs the actions and intent probabilities at time $T_{horizon}$. The softmax probabilities from each task are then used to calculate the specific action/behavior loss value $L_{act}^n = \sum_{i=1}^{K} CE(class_{act}^n, \hat{class_{act}^n})$, where $act_n$ is the activity, $K$ are the maximum number of VRU instances from the sequence and $class_{act}^n, \hat{class_{act}^n}$ are the ground truth and predicted class labels respectively. Using the following weighted sum of individual cross-entropy losses, we jointly train all the classification tasks:

\begin{equation}
    L_{action} = \omega_1 L_{gait} + \omega_2 L_{attn} + \omega_3 L_{ornt} + \omega_4 L_{dist} + \omega_5 L_{crossing} 
\end{equation}

For trajectory prediction we use a LSTM encoder-decoder configuration focusing on the bounding box inputs. The encoded inputs are passed to the LSTM decoder stack along with the internal state and output future box center positions in pixels. Given the bounding box sequences from time 1 to $T_{current}$, the model predicts the future bounding box centers from time $T_{current}$ to $T_{horizon}$. This regression task is trained to optimize mean square error(MSE) loss function $L_{MSE} = \frac{1}{N}\sum_{i=T_{current}}^{T_{current+horizon}} (I[x, y] - \hat{I[x, y]})^2$. Here $I and \hat{I}$ are the ground truth and predicted pixel values of the object box centers respectively. We added a L2 regularization term to avoid overfitting $L_{traj} = L_{MSE}(I, \hat{I}) + \lambda_{reg} L2_{reg}$.
The total loss function that is jointly optimized as weighted sum of classification and regression losses $L_{act} and L_{traj}$ respectively.

\section{Experiments and Results}\label{sec:Experiments} 

\subsection{Dataset Processing and Augmentation}
JAAD dataset has 2D bounding boxes and tracks annotated for persons in the scene and only a unique set of pedestrians come with behavior labels and respective contextual information. For our training we mainly concentrate on the pedestrians that are not heavily occluded (less than 25\% visible) and with crossing intent and behavior annotations (less than 30\%). This gives us very few sequences ($\sim$40K frames) and also has imbalance in class labels. The number of pedestrians that are actually seen to be distracted or phoning are almost less than 20\% of all behavior labels. Firstly, we extract the tracks for pedestrians, with durations greater than 1.5 seconds. Using a sliding window we process the sequences to generate tracks with observation lengths of 0.5s and 1s i.e. 15 and 30 frames (given 30Hz sequences), and prediction lengths of 1s i.e. 30 frames. We pad some tracks with shorter sequence lengths at the beginning and end of the tracks in order to have a fixed sequence length for the entire dataset.

The generated tracks are then processed through the perception pipeline to extract 2D pose and semantic segmentation masks. Additional training tracks are generated from augmentation by flipping the sequences, adding pixel dropout and random noise to scene masks and pose keypoints.

\subsection{Training Implementation}
In this section we describe how we train the baseline models for individual action recognition tasks using pre-computes pose features. We present the network architecture for the multi-task model using \textit{VRUNet}, and then compare our results to the baseline models and JAAD benchmark \cite{Rasouli_2017_ICCV}.

\textbf{\textit{Modular Approach - Baseline Training:}} To classify gait (Walking or Standing), a Resnet10 model with 1D convolutions is trained using input features of length $N\times6$ (from section \ref{subsec:models}) and outputs class probability. Model is trained to optimize a binary cross-entropy loss, for 100 epochs using an initial learning rate (\textit{lr}) of 0.0001 and adam optimizer for a batch size of 32. For attention and orientation we use the same model considering that these are mostly determined by common features of the person's face and shoulders. A Resnet10 model with 1D convolutions is jointly trained for these two different tasks, using a weighted cross-entropy loss. The model is trained for 150 epochs with adam optimizer. The model outputs probabilities for attention and orientation, and jointly classifies if the person is Looking or not and if so in what direction (Left, Right, Front, Back) is he relatively oriented. A distraction model is fitted with a SVM classifier using averaged pose features for a sequence length of N (see section \ref{subsec:models}). 

These models for individual tasks are used to generate action predictions on the tracks for crossing intent. Now given the actions for a VRU for the entire track history and the scene context (see section \ref{subsec:models}) we classify if the pedestrian is Crossing or not. This is trained with the ground truth label from the last sample of the prediction length of track (1 second horizon).

\textbf{\textit{VRUNet Training:}} We use 60\%, 20\% ,20\% splits for training, validation and test sequences from all the tracks. Inputs to the model are 2D pose and bounding box sequences for each person in the scene and also the scene mask as shown in \ref{tab:VRUNet-input}. Pose and bounding boxes are normalized with input resolution. Segmentation mask output is binary encoded for each class (classes = 5). Input sequences of object and scene context for each person are for the observed length (1 to $T_{current}$). For action recognition (gait, attention and distraction) and orientation we use the training labels from current time ($T_{current}$). For crossing intent the training label is the state at end of prediction horizon ($T_{current+horizon}$). We use a stateless LSTM for encoding pose and bounding box features and pass the state of the encoder to the decoder for the trajectory branch. For trajectory training we use the future horizon bounding box centers as training labels from time $T_{current}$ to $T_{current+horizon}$. Hyperparameters used for training are as shown in Table \ref{tab:VRUNet-hyperparameters}.

\begin{table}[h]
\begin{center}
\caption{Training Hyperparameters}
\label{tab:VRUNet-hyperparameters}
\begin{tabular}{r|c}
Name & Value \\
\hline
Batch Size & 32 \\
Learning Rate & $10^{-5}$ with adaptive step change \\
Optimizer & Adam \\
Epochs & 500 \\
Regularization & L2 for Trajectory Outputs 0.0003 \\
Activation & LSTM with tanh and Conv2D with relu \\
\end{tabular}
\end{center}
\end{table}

\subsection{Results}
In Table \ref{tab:Accuracy Comparison} we have the comparison for average precision (AP) of the action recognition and crossing intent prediction for a 1 second horizon. We present the results for the modular approach i.e. separately trained models for two observation durations of 0.5 and 1 seconds and for the VRUNet multi-task model with observation duration of 1 second or 30 frames. 

\definecolor{green}{rgb}{0,0.5,0}
\begin{table}[h]
\begin{center}
\caption{Average Precision (AP\%) for behaviors and intent predictions (GAIT - gait, ATTN - attention, DIST - distraction, ORNT - orientation, XNG - crossing intent)(Red - Ours)}
\label{tab:Accuracy Comparison}
\begin{tabular}{l|c|c|c|c|c}
Models &  GAIT & ATTN & DIST & ORNT & XNG \\
\hline
AlexNet & 78.34 & 67.45 & N/A & N/A &N/A \\
\hline
AlexNet-Pre & 80.45 & 75.23 & N/A & N/A & N/A \\
\hline
AlexNet-Crop & 83.45 & 80.23 &  N/A & N/A & N/A \\
\hline
Context & N/A & N/A &  N/A & N/A & 62.73 \\ 
\hline
\textbf{\textcolor{red}{Modular (0.5s)}} & 91.32 & 87.5 &  86.45 & 83.52 & 66.87 \\
\hline
\textbf{\textcolor{red}{Modular (1s)}} & 93.27 & 89.35 &  87.43 & 83.27 & 67.34 \\
\hline
\textbf{\textcolor{red}{VRUNet (1s)}} & 79.35 & 78.27 &  81.23 & 82.37 & 73.47 \\
\end{tabular}
\end{center}
\end{table}

\begin{figure}[h]
    \centering
    \includegraphics[width=\linewidth, height=4cm]{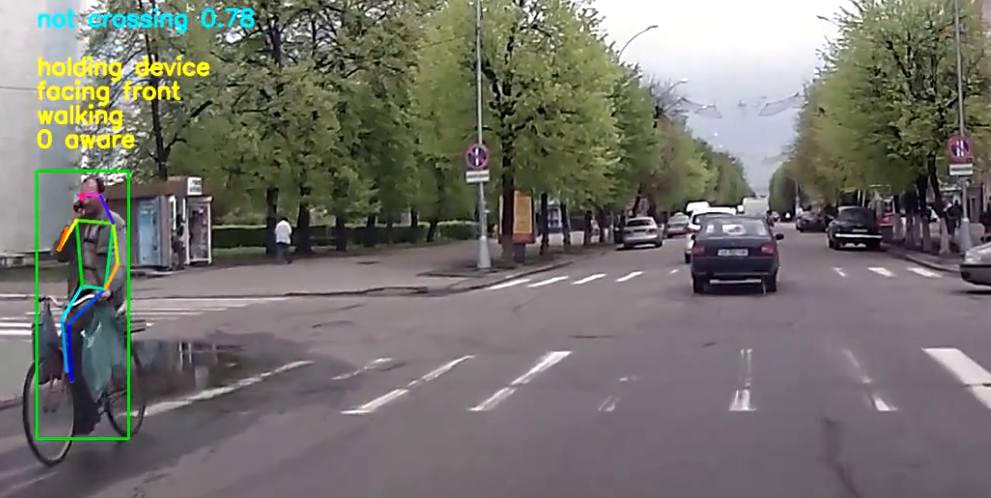}
    \caption{Distracted cyclist not crossing the road with the class labels from VRUNet prediction (Not crossing, phoning, facing front, walking, aware)}
    \label{fig:not crossing}
\end{figure}

\begin{figure*}[h]
\centering
\begin{subfigure}{0.95\textwidth}
    \includegraphics[width=\textwidth, height=4cm]{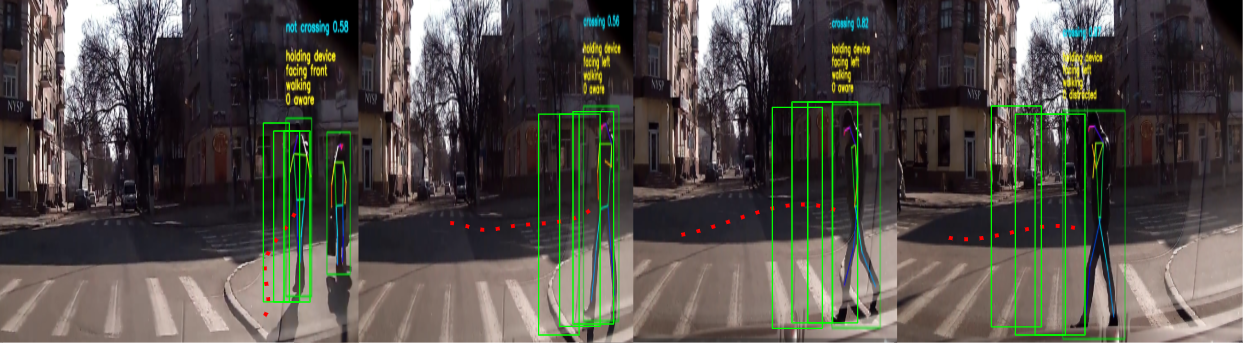}
    \caption{crossing sequence prediction of a distracted crossing pedestrian. Sequence start to end (left to right), model starts to predict the intent as crossing from second frame instance. Red color trajectory is the smooth and fitted prediction from multi-task model.Green boxes are the actual ground truth values for horizon of 1s.}
    \label{fig:vrunet_a}
\end{subfigure}%
\newline
\newline
\newline
\begin{subfigure}{0.95\textwidth}
    \includegraphics[width=\textwidth]{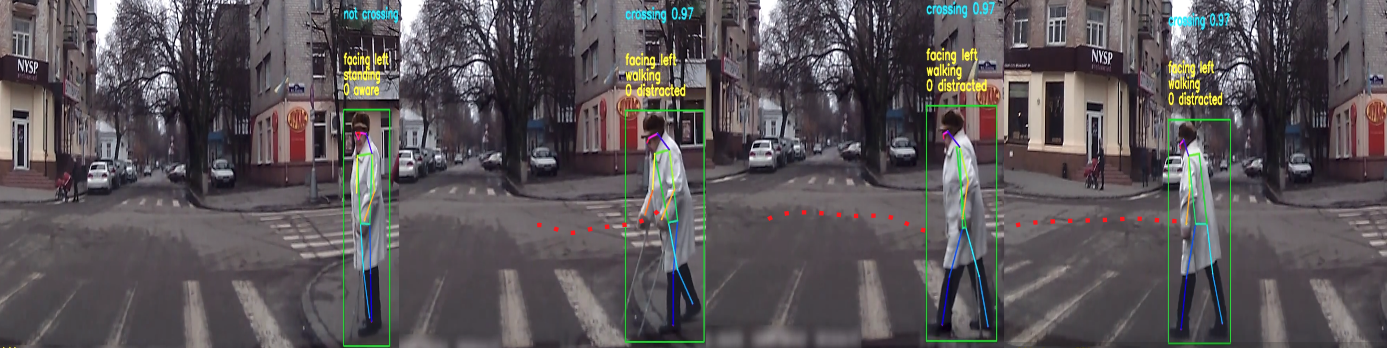}
    \caption{crossing sequence prediction of pedestrian.}
    \label{fig:vrunet_b}
\end{subfigure}
\newline
\newline
\caption{Results visualization of VRUNet prediction}
\label{fig:vrunet}
\end{figure*}

We see that the overall AP of individual action recognition tasks is much better when we train for each task separately. The crossing intent prediction for a 1 second horizon is better when the history or observation length is higher (1s). Using the 2D pose for persons and their features in separate models (modular), we outperform the accuracy for gait (Walking/Standing) and attention (Looking/Not Looking). Also adding some high level scene context to the crossing intent prediction task we see that the accuracy is higher than the baseline from JAAD authors. Using our multi-task VRUNet model, action recognition AP\% is lower compared to modular approach, as we avoid over-fitting for any specific task with weight sharing and weighted loss function. We see that we outperform the overall crossing intent accuracy, and this shows the benefit of including low level features from scene masks as input features for training. For trajectory prediction we currently do not have a baseline comparison to any models. The predicted trajectory is filtered and we fit a third order polynomial. We then generate trajectory points from the polynomial using pixel positions. Qualitative results for action recognition, intent prediction and trajectory are shown in Figures \ref{fig:not crossing} and \ref{fig:vrunet}. In Figure \ref{fig:vrunet_a} we see the predicted trajectory points in red and the actual sampled ground truth boxes for the pedestrian for 1s horizon are shown in green.

\section{Conclusions}\label{sec:Conclusions} 
In this paper, we have presented a multi-task learning model to jointly predict the actions, crossing intent and also the trajectories for VRU's. We have developed the model, by using low level pedestrian context i.e. 2D human pose and object bounding boxes, and scene context in the form of segmentation masks. We separately encode the visual features for the sequence of VRU poses, boxes and segmentation masks and finally fuse them to predict action states (gait, attention, distraction), orientation and crossing intent of VRU's. We also simultaneously predict the trajectory or future positions in images, using the encoded box features. We showed the improved performance of action prediction and crossing intent using this approach on the JAAD dataset. In addition we have trained separate models for each task to establish a benchmark and compare the results with the multi-tasking approach. We specifically see that the model generalizes well for all the action recognition tasks and crossing intent prediction improves using the scene context. 

In future research we could achieve better accuracy for action recognition tasks and also increase the prediction horizon for crossing intent with improved datasets and better model architectures. Vehicle odometry information as input could improve the accuracy for trajectory prediction and overall intent prediction. Our future work will focus on including social interaction between VRU's, and interactions between the person and other objects in the scene to further improve the quality of predictions. Our work has not yet been validated for different populations which is very important to be able to have such behavioral intelligence for automated driving. 



{\small
\bibliographystyle{IEEEtran}
\bibliography{references}
}


\begin{biography}

\noindent \textbf{Adithya Ranga} is a lead machine learning researcher for Valeo driving assistance research team. His main research focus is machine learning, with applications in computer vision. He is currently dedicated towards developing advanced perception, behavior prediction and motion planning algorithms for autonomous driving. Prior to joining Valeo, he received his Masters Degree (majoring in control systems, dynamics and machine learning) from University of Michigan - Ann Arbor (2012-13) and worked for more than 4 years at ford motor company as a software researcher for autonomous systems.  \\

\noindent \textbf{Filippo Giruzzi} is a student at MINES ParisTech, studying for a Master of Science and Executive Engineering, with a major in Applied Mathematics. Part of this work was completed during his Software Engineer internship within the Driving Assistance Research team at Valeo North America. His main interests and work experience are in Deep Learning and Computer Vision, with applications in predictive modeling and behavior prediction for autonomous driving. \\

\noindent \textbf{Jagdish Bhanushali} is Deep learning researcher and Software Engineer working at Valeo in driving assistance research team. His main focus is applying deep learning and computer vision algorithms on autonomous driving. He started with Valeo as an intern and continued to grow his career in Deep learning software engineer. Before working at Valeo, he completed his Masters in Computer Science from Santa Clara University during 2016-2018. \\

\noindent \textbf{Emilie Wirbel} is Software Team Leader for Deep Learning and Deep Learning expert at Valeo Driving Assistance Research and research scientist at Valeo.ai France. She received an engineering master degree from the French engineering school Mines ParisTech (2008-2011), then a PhD in robotics in 2014, and has been working at Valeo since 2015. Her work focuses on Deep Learning for prediction and control in the autonomous driving context, in particular relying on end-to-end imitation learning. \\

\noindent \textbf{Tuan-Hung Vu} is a Research scientist at valeo.ai, France. He received his PhD from \'Ecole Normale Sup\'erieure, under the supervision of Ivan Laptev. He completed a ``Master 2'' Degree in Mathematics, Machine Learning and Computer Vision (MVA) in \'Ecole Normale Sup\'erieure de Cachan. He obtained the engineering degree from Télécom ParisTech in 2014.

\end{biography}

\end{document}